\PassOptionsToPackage{table}{xcolor}
\documentclass{article}

\usepackage[preprint]{neurips_2025_db}

\usepackage[utf8]{inputenc} 
\usepackage[T1]{fontenc}    
\usepackage{hyperref}       
\usepackage{url}            
\usepackage{booktabs}       
\usepackage{amsfonts}       
\usepackage{nicefrac}       
\usepackage{microtype}      
\usepackage{xcolor}         

\definecolor{headerblue}{rgb}{0.8, 0.85, 1}
\hypersetup{
    colorlinks=true,
    linkcolor=steelblue,
    urlcolor=steelblue,
    citecolor=steelblue
}
\definecolor{steelblue}{RGB}{70,130,180}

\usepackage{graphicx}
\usepackage{booktabs}
\usepackage{multirow}
\usepackage{makecell}

\usepackage{fontawesome5}
\usepackage{adjustbox}
\usepackage{ulem}
\usepackage{amsmath}
\usepackage{enumitem}
\usepackage{pifont}
\newcommand{\cmark}{\ding{51}}  
\newcommand{\xmark}{\ding{55}}  
\usepackage{caption}

\title{X2C: A Large-Scale Benchmark for Nuanced \\Humanoid Facial Expression Imitation}

\author{%
  Peizhen Li$^{1}$,
  Longbing Cao$^{1}$, 
  Xiao-Ming Wu$^{2}$,
  Runze Yang$^{1}$,
  Xiaohan Yu $^{1}$\\
  $^{1}$Macquarie University
  $^{2}$ Nanyang Technological University
}

\begin{document}

\maketitle

\begin{abstract}

Fine-grained facial expression transfer from humans to humanoid agents presents a unique pattern recognition challenge due to the significant domain gap between biological facial dynamics and mechanical control spaces. While visual synthesis of talking heads has advanced rapidly, mapping high-dimensional visual cues to precise, physically constrained actuation signals remains an open problem, primarily due to the lack of large-scale paired data. To bridge this gap, we introduce \textbf{X2C}, a comprehensive benchmark dataset comprising 100,000 $\langle \text{image}, \text{control value} \rangle$ pairs. Unlike existing resources, X2C features nuanced, physically grounded expressions annotated with 30 continuous control parameters, establishing a high-fidelity standard for this task. Building on this resource, we propose \textbf{X2CNet}, a two-stage deep learning framework that explicitly decouples visual motion features from mechanical control regression to model the correspondence between human perceptual cues and humanoid actuation. Extensive experiments, including quantitative benchmarking and real-world physical validation, demonstrate that our approach achieves superior cross-domain consistency and enables robust, in-the-wild expression imitation. 

\faGithub\ \textbf{Code \& Data}: \href{https://lipzh5.github.io/X2CNet/}{https://lipzh5.github.io/X2CNet/} 
\end{abstract}

\section{Introduction}
Rapid progress in humanoid robotics has been observed across various domains such as reception~\cite{stock2016emotion}, education~\cite{kanda2004interactive, tanaka2007socialization} and healthcare~\cite{kozima2005interactive, esubalew2012step}, where humanoid robots engage in affective communication with humans. These robots are increasingly deployed to interact socially, provide assistance, or support learning, making their ability to express emotions a key factor in fostering trust, empathy, and engagement~\cite{rawal2022facial, ottoni2024systematic, jung2021users, seyitouglu2024robots, saunderson2019robots}. Effectively delivering affective information is crucial for enhancing user experiences and ensuring meaningful interactions between humans and robots. 
Consequently, there has been growing emphasis on enabling humanoid robots to imitate realistic and authentic facial expressions, as facial expressions play a central role in conveying emotional cues~\cite{mehrabian2017communication,li2024ugotme}. 

Despite recent advances in humanoid facial expression imitation~\cite{chen2021smile, hu2024human}, the fidelity of humanoid facial expressions---especially the fine-grained, nuanced emotional cues---remains difficult to guarantee due to the scarcity of data required to learn emotional subtleties and guide informed on-robot execution. 
Existing datasets of humanoid facial expressions for the imitation task (Smile~\cite{chen2021smile}, Coexpression~\cite{hu2024human}) are typically small in size, lack sufficient data diversity (e.g., they do not include asymmetric facial expressions), and the emotional nuances that can be learned are limited by low annotation dimensionality (see Table~\ref{tab:benchmark-stat}).
Their annotation accuracy is not guaranteed as the dataset collection relies on facial landmark predictions~\cite{baltruvsaitis2016openface}, which introduce prediction errors.


\begin{figure*}
    \centering
    \includegraphics[width=1.0 \linewidth]{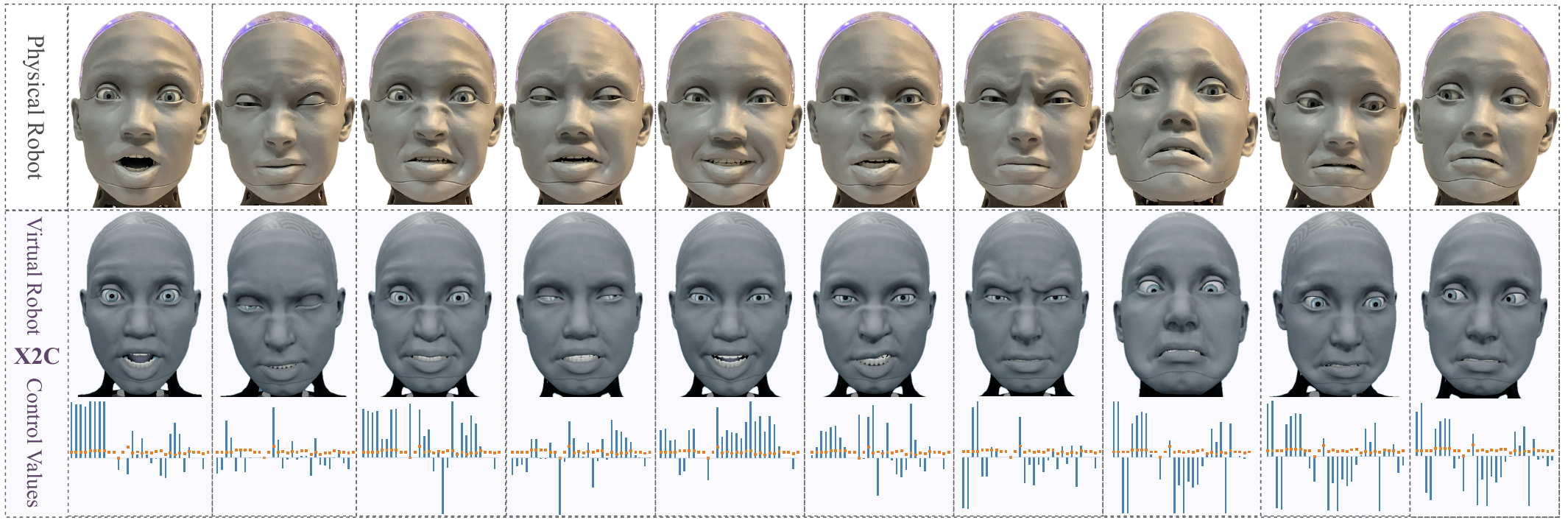}
    \caption{\textbf{Demonstration of X2C dataset examples.} Each example in the X2C dataset consists of: (1) an image depicting the virtual robot, shown in the middle; and (2) the corresponding control values, visualized at the bottom. In these visualizations, the height of each \textbf{blue bar} represents the magnitude of the corresponding value, while the \textbf{orange dots} indicate the values in the neutral state.
}
\label{fig:virtual_physical}
\end{figure*}
\begin{table}[h]
\centering
\caption{Summary statistics of existing datasets for realistic humanoid facial expression imitation.}
\label{tab:benchmark-stat}
\resizebox{\textwidth}{!}{%
\begin{tabular}{llcccccccc}
\toprule
\multirow{3}{*}{\makecell{~~~~~\textbf{Dataset}}}  & \multicolumn{4}{c}{\cellcolor{headerblue}\textbf{Dataset Characteristics}} & \multicolumn{3}{c}{\cellcolor{headerblue}\textbf{Dataset Quality}}  \\
\cmidrule(lr){2-5} \cmidrule(lr){6-8} \cmidrule(lr){9-10}
& \makecell{Size} & \makecell{Asymmetric\\Expressions}& \makecell{Input \\ Dimensionality}  & \makecell{Annotation \\ Dimensionality} & \makecell{Annotation \\ Accuracy}& \makecell{Data \\ Alignment} & \makecell{Data \\Diversity}  \\
\midrule
\makecell{\textbf{X2C}}  &\textbf{100,000} & \cmark &$512\times512\times3$   & \textbf{30} &  \ding{72}\ding{72}\ding{72}\ding{72}\ding{72} & \cmark & \ding{72}\ding{72}\ding{72}\ding{72}\ding{72} \\
\makecell{Smile~\cite{chen2021smile}} &15,000 & \xmark & $480\times320\times3$ & 10 & \ding{72}\ding{72}\ding{72}\ding{72} & \cmark & \ding{72}\ding{72}\ding{72} \\
\makecell{Coexpression~\cite{hu2024human}} &1000 & \xmark & $113\times2$ & 11 & \ding{72}\ding{72}\ding{72}\ding{72} & \cmark & \ding{72}\ding{72}\ding{72} \\

\bottomrule
\end{tabular}%
}
\end{table}

To bridge the gap, we introduce \textbf{X2C}, a new resource for realistic humanoid facial expression imitation. 
It consists of 100,000 (image, control value) pairs, with each image depicting a humanoid robot displaying a diverse range of nuanced facial expressions (Figure~\ref{fig:virtual_physical}). Each image is annotated with 30 numerical control values representing the ground-truth expression configuration. These control values can be used to drive the physical humanoid robot to reproduce the expression shown in the image.
\textbf{Asymmetric facial expressions}~\cite{rinn1984neuropsychology} are also included in our dataset (e.g., the 2nd column in Figure~\ref{fig:virtual_physical}) to simulate the human-like behavior and encourage diversity. 
To facilitate understanding of the relationship between the physical and virtual robots, images of the {physical robot} are also included at the top of Figure~\ref{fig:virtual_physical}. 
The {physical robot} and its {virtual counterpart} share the same set of controls, {ensuring consistent facial expressions across both platforms. In principle, facial expressions should \textbf{be independent of} the skin type (or identity) of the humanoid robot~\cite{liu2024norface, zhang2021learning}. Therefore, issues such as sim-to-real gap~\cite{peng2018sim}} do not raise, since control values encode emotional nuances and serve as a bridge between the virtual and physical robots. 

Images in the dataset are extracted from videos of the humanoid robot performing facial expression animations. These animations are manually curated by volunteers from different birth countries and of different genders (including females and males), to help eliminate potential biases related to cultural background and gender.
The dataset includes basic facial expressions (e.g., surprise, joy, and sadness~
\cite{ekman1992argument}) at varying intensities (e.g., the last two in Figure~\ref{fig:virtual_physical} show fear with subtle differences in gaze and head pose), as well as complex expressions that may not fit neatly into basic emotion categories (e.g., the 4th column in Figure~\ref{fig:virtual_physical}). This is done purposefully to ensure broad \textbf{expression coverage}.  
To ensure \textbf{consistency}, the upper-body pose of the robot remains fixed during video recording, and all images are resized to a uniform resolution. To guarantee the \textbf{uniqueness} of the dataset, we apply structural similarity checks to the captured images to identify and remove near-duplicate frames, resulting in 100,000 retained images with diverse humanoid facial expressions for annotation.
To ensure annotation \textbf{accuracy}, we formulate interpolation equations based on parameters specified in the animation files. For control value sampling, i.e., given a timestamp $s$, precise continuous control values will be provided by the equations. The timestep for sampling both control values and images is set to 0.05 seconds, enabling perfect alignment between images and their corresponding control annotations~\cite{northcutt2021confident,koh2021wilds,northcutt2021pervasive}, thereby ensuring the dataset quality. 
Details of the control values sampling and annotation process are provided in Section~\ref{sec:dataset-sample-annotation}.
Dataset characteristics such as size (number of samples), input dimensionality, and annotation dimensionality, as well as dataset quality metrics such as annotation accuracy, data alignment and data diversity~\cite{paullada2021data} are summarized in Table~\ref{tab:benchmark-stat}.
To our knowledge, \textbf{X2C} is the first high-quality, high-diversity, large-scale dataset featuring nuanced humanoid facial expressions specifically designed for realistic humanoid imitation. Equipped with \textbf{X2C}, we introduce \textbf{X2CNet}, a novel framework for realistic humanoid facial expression imitation. It decomposes the humanoid learning process into two stages: in the first stage, the expression dynamics is captured through a motion transfer module; in the second stage, the correspondence between nuanced humanoid facial expression and their underlying control values is learned via large-scale training using \textbf{X2C}.
This framework enables facial expression imitation in the wild for different human performers. An overview of \textbf{X2CNet} is provided in Figure~\ref{fig:framework}. 

After introducing \textbf{X2C} (Section~\ref{sec:dataset}), we demonstrate its value in enabling human-to-humanoid learning by presenting an imitation framework, \textbf{X2CNet} (Section~\ref{sec:framework}), and further validate its potential through real-world experiments on physical humanoid robots performing nuanced facial expression imitation tasks.
Beyond imitation, \textbf{X2C} opens up new research avenues, including human-like motion generation, facial expression evaluation for robots~\cite{gielniak2013generating, becker2011evaluating} and the development of expressive humanoid robots for affective human–robot interaction~\cite{breazeal2003emotion}.

\section{The X2C Dataset}
\label{sec:dataset}
\begin{figure}
    \centering
\includegraphics[width=1.0\linewidth]{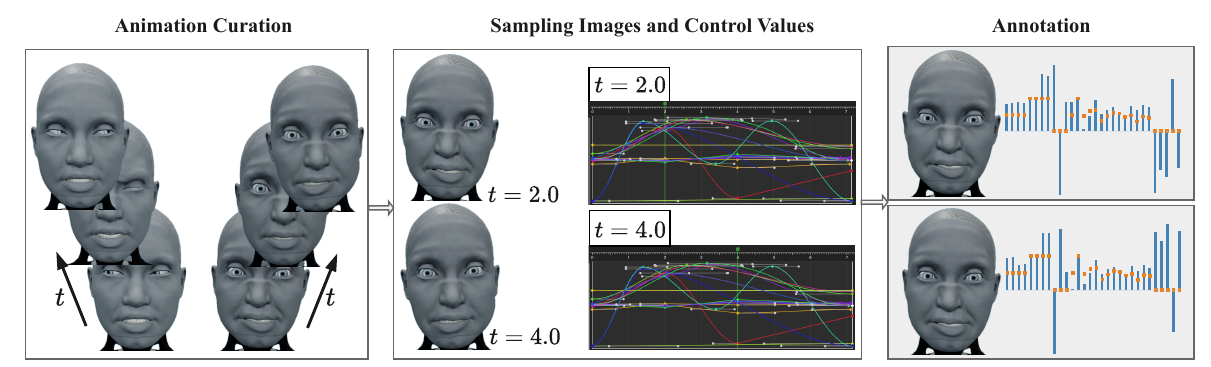}
    \caption{\textbf{The pipeline for dataset collection}. We first curate humanoid facial expression animations covering all basic emotions and beyond. Images and their corresponding control values are then sampled at the same timestamps (e.g., if an image is sampled at $t=2.0$, its control value annotation is also sampled at $t=2.0$) to obtain the temporally aligned pairs. See Section~\ref{sec:dataset-sample-annotation} for more details.}
    \label{fig:dataset_pipeline}
\end{figure}
\textbf{X2C} has been made publicly available and provides images of nuanced humanoid facial expressions along with ground-truth control value annotations. A comparison of the data characteristics and quality of \textbf{X2C} against existing datasets for realistic humanoid facial expression imitation is presented in Table~\ref{tab:benchmark-stat}.

In the following sections, we first introduce the humanoid robot and the control value preliminaries (Section~\ref{sec:humanoid-intro}).
Before providing an overview of the dataset (Section~\ref{sec:dataset-characteristics}).
we detail the dataset collection process:
volunteers were invited to curate facial expression animations for the humanoid robot and record videos (Section~\ref{sec:dataset-video-filming}); we then sampled images from videos and calculated control values using mathematical equations (Section~\ref{sec:dataset-sample-annotation}) to construct the (image, control value) pairs.
The pipeline of dataset collection is provided in Figure~\ref{fig:dataset_pipeline}. 
10 volunteers were recruited from the student population at Macquarie University, all of whom were over 18 years old and provided informed consent. 
To minimize possible biases related to different background, we purposefully selected volunteers with different birth countries, including undergraduate and PhD students of different genders. Dataset collection ran from 15nd November 2024 to 15nd January 2025\footnote{Ethics approval, data collection, and analysis was led by researchers from Macquarie University.} 


\subsection{The Humanoid Robot and Control Values }
\label{sec:humanoid-intro}
\begin{figure}[h]
    \centering
\includegraphics[width=0.6\linewidth]{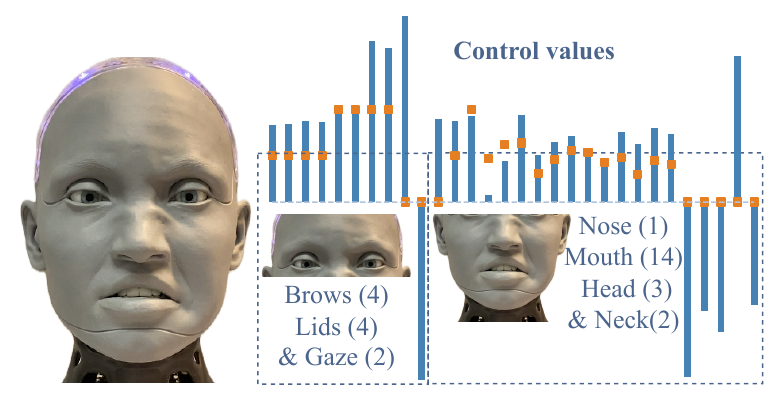}
    \caption{\textbf{An illustration of the correspondence between control values and control units.} In the control value visualization, the first 4 values control the brow movements, the next 4 control eyelid motions, and so on for the other units.}
    \label{fig:exp-ctrl}
\end{figure}
The humanoid robot employed for dataset collection is called Ameca (Figure~\ref{fig:ameca}), which features 32 Degrees of Freedom (DoFs) including facial actuators and head/neck movements ({Figure~\ref{fig:32DoF}). The higher number of facial DoFs---compared with most existing humanoid robots~\cite{zhang2025fabg,li2023design,faraj2021facially,kerzel2017nico})---allows for finer-grained and more nuanced facial expressions. 
There are 30 control values associated with these DoFs, which are responsible for driving actuators located at different expression-relevant control units, including the brows, lids, gaze, nose, mouth, head and neck. 

The distribution of control values across these units is visualized in Figure~\ref{fig:exp-ctrl}, and the value ranges for each control are provided in Table~\ref{tab:ctrl_range}.
Note that the number of control values and DoFs differ. This is because the gaze controls (Gaze Target Phi and Gaze Target Theta) drive the left and right eyes symmetrically, which reflects a human-like design, as humans exhibit conjugate gaze~\cite{martinez2004role}.
\begin{table*}[t]
\centering
\caption{Names of the controls and the corresponding ranges of their values. }
\begin{adjustbox}{width=\columnwidth,center}
\begin{tabular}
{cccccc}
\toprule
Jaw Pitch& Jaw Yaw& Lip Bottom Curl  &Lip Bottom Depress Left  &Lip Bottom Depress Middle & Lip Bottom Depress Right   \\
$[0, 1]$ &$[0, 1]$  &$[0, 1]$  &$[0, 1]$  &$[0, 1]$   &$[0, 1]$ \\
\midrule
Lip Corner Raise Left & Lip Corner Raise Right& Lip Corner Stretch Left  &Lip Corner Stretch Right  &Lip Top Curl & Lip Top Raise Left    \\ 
$[0, 1]$ &$[0, 1]$  &$[0, 1]$  &$[0, 1]$  &$[0, 1]$   &$[0, 1]$     \\ 
\midrule
Lip Top Raise Middle& Lip Top Raise Right  & Nose Wrinkle &Brow Inner Left & Brow Inner Right  & Brow Outer Left\\ 
$[0, 1]$&  $[0, 1]$ &$[0, 1]$ &$[0, 1]$ &$[0, 1]$  &$[0, 1]$     \\ 
\midrule 
Brow Outer Right& Eyelid Lower Left  & Eyelid Lower Right &Eyelid Upper Left & Eyelid Upper Right  & Gaze Target Phi\\ 
$[0, 1]$&  $[-1, 2]$ &$[-1, 2]$ &$[-1, 2]$ &$[-1, 2]$  &$[-2.3, 2.3]$     \\ 
\midrule 
Gaze Target Theta& Head Pitch  & Head Roll &Head Yaw & Neck Pitch  & Neck Roll\\ 
$[-1.1, 1.1]$&  $[-0.5, 0.3]$ &$[-0.3, 0.3]$ &$[-0.5, 0.5]$ &$[-0.3, 0.5]$  &$[-0.3, 0.3]$     \\ 
\bottomrule
\end{tabular}
\end{adjustbox}
\label{tab:ctrl_range}
\end{table*}

\subsection{Humanoid Expression Animations}
\label{sec:dataset-video-filming}
\paragraph{Environment Preparation} Exploration of the robot's expression space requires arbitrary combinations of different control values (each sampled within its legal range), which is necessary to curate a high-diversity dataset. Some combinations may lead to rare facial expressions that most humans cannot display (Figure~\ref{fig:weird}), and frequently driving the robot with such control values can cause irreversible damage to the robot's silicon skin, leading to expensive repairs.
For safety and to minimize mechanical wear on the physical robot, we chose to conduct dataset collection in a simulation environment where a virtual counterpart of the physical robot is available. Given the same control values, the virtual robot displays the same facial expressions as the physical one. There will be no issues such as the sim-to-real gap~\cite{peng2018sim} since facial expressions should be disentangled from the robot’s skin (or identity)~\cite{liu2024norface, zhang2021learning}. {Images from the physical robot and its virtual counterpart are equivalent in the sense that they have the expression embedding in robot's action space, represented by control values.}
Details of the data collection environment are provided in the Supplemental Material. This environment can be accessed simultaneously by multiple certified accounts, which accelerates the data collection process.
\paragraph{Expression Animation Curation} After 30 hours of training, volunteers became familiar with rigging the robot in the aforementioned environment and acquired the necessary prerequisites for animators. They were then tasked with creating key-framed animations~\cite{safonova2004synthesizing,liu2002synthesis}, each defined by a sequence of critical frames that capture the most significant humanoid expressions at key moments in time, along with corresponding interpolation methods~\cite{parent2012computer}. Intermediate frames (in-betweens) are then interpolated to produce smooth expression animations. The currently available interpolation methods include \textbf{Cubic Bézier}, \textbf{Linear} and \textbf{Step} interpolations. 
\textbf{Cubic Bézier} interpolation provides a smooth transition by blending four control points \( P_0, P_1, P_2, P_3 \), where \( P_0 \) and \( P_3 \) are the start and end points, and \( P_1 \), \( P_2 \) are control points. The interpolation is defined over the normalized parameter \( u \in [0, 1] \)~\cite{bezier1972numerical}:
\begin{equation}
\label{eq:bezier}
   I(u) = (1-u)^3P_0 + 3(1-u)^2uP_1 + 3(1-u)u^2P_2 + u^3P_3, ~~u\in [0, 1].
\end{equation}
\textbf{Linear} interpolation creates a straight-line transition between two keyframe values \( P_0 \) and \( P_1 \) defined at times \( t_0 \) and \( t_1 \), respectively. The interpolation function is given by~\cite{foley1996computer}:
\begin{equation}
    \label{eq:linear}
I(t) = P_0 + (P_1 - P_0) \cdot \frac{t - t_0}{t_1 - t_0}, \quad t \in [t_0, t_1].
\end{equation}
\textbf{Step} interpolation holds a constant value until the next keyframe. For two keyframes at times \( t_0 \) and \( t_1 \), with values \( P_0 \) and \( P_1 \), the step interpolation is defined as~\cite{alan1992advanced}:
\begin{equation}
    \label{eq:step}
    I(t) =
\begin{cases}
P_0, & \text{if } t \in [t_0, t_1) \\
P_1, & \text{if } t = t_1
\end{cases}
\end{equation}
Volunteers could specify the interpolation method and value for each control at a critical moment to form keyframes. These values were purposefully selected to sweep the full range of each control as much as possible, ensuring broad coverage of the expression space and promoting diversity in the dataset. The resulting 560 animations (with durations ranging from 1 to 15 seconds) then underwent a subsequent processing stage.

\subsection{Sampling and Annotation}
\label{sec:dataset-sample-annotation}
We filmed videos of all curated animations using the same device (MacBook Air, 2020) and under consistent configurations (i.e., frame rate, and upper-body pose of the robot) to ensure data consistency. Images were then sampled at a constant timestep of $s = 0.05$ seconds and resized to a uniform resolution of 512$\times$512 pixels. To ensure data uniqueness, the resulting images were processed with a structural similarity check to identify and remove near-duplicate frames if their similarity exceeded a threshold ($\theta = 0.99$). Given two image patches $x$ and $y$, the Structural Similarity Index (SSIM) is computed as follows~\cite{wang2004image}:
\begin{equation}
    \text{SSIM}(x, y) =
\frac{(2\mu_x \mu_y + C_1)(2\sigma_{xy} + C_2)}
     {(\mu_x^2 + \mu_y^2 + C_1)(\sigma_x^2 + \sigma_y^2 + C_2)}, 
\end{equation}
where $C_1$ and $C_2$ are stability constants, defined as $C_1 = (K_1 L)^2$ and $C_2 = (K_2 L)^2$, with $L=255$, $K_1=0.01$ and $K_2=0.03$. $\mu_x$, $\mu_y$ denote the mean of $x$ and $y$ respectively; $\sigma^2_x$ and $\sigma^2_y$ are their variances; and $\sigma_{xy}$ is the covariance between $x$ and $y$. 
To obtain the control value annotations, we retrieve the keyframes and interpolation parameters from the animation metadata, formulate the corresponding interpolation equations (as shown in equations~\eqref{eq:bezier} to~\eqref{eq:step}) and sample precise control values at the same timestamps used to extract images from the animations. This ensures annotation accuracy and temporal alignment between the images and their annotations. 

The overall dataset collection pipeline is summarized in Figure~\ref{fig:dataset_pipeline}, where we provide a visualization for control curve sampling at two timestamps (i.e., $t=2.0$ and $t=4.0$).

\subsection{Dataset Overview}
\label{sec:dataset-characteristics}

\begin{table*}[h]
\centering
\caption{Summary statistics of the 30 control values.
The names of the controls are abbreviated using the first capital letter of each word. Full names are provided in Table~\ref{tab:ctrl_range}.}
\begin{adjustbox}{width=\columnwidth,center}
\begin{tabular}
{cccccccccccccccc}
\toprule
& JP& JY &  LBC &LBDL & LBDM &LBDR&LCRL&LCRR&LCSL& LCSR&LTC& LTRL&LTRM&LTRR& NW \\ 
\midrule
$\mu$&  0.635 &0.519 &0.544 &0.565 &0.465 &0.560 &0.355 &0.542 &0.646 &0.403 &0.543 &0.649 &0.540 &0.620 &0.191      \\ 
$\sigma$&0.366 &0.190 &0.123 &0.085 &0.087 &0.064 &0.129 &0.087 &0.119 &0.095 &0.153 & 0.134 &0.182 &0.140 &0.287 \\
$V_{\text{max}}$&1.000 &1.000 & 0.991 & 0.750 & 0.759 &0.822 &0.802 &0.976 &1.000 & 0.991 & 1.000 & 1.000  & 1.000 & 1.000  &1.000 \\
$V_{\text{min}}$&0.000 & 0.000 & 0.000 &0.148 &0.000 &0.185 & 0.000 &0.250&0.295&0.300&0.000&0.475&0.300 &0.409 &0.000 \\
${V}_{\text{neu}}$&1.000 &0.500 &0.460 &0.560 &0.430 &0.540 & 0.470 & 0.620 & 0.640 &0.310 &0.410 & 0.480 &0.300 &0.450 &0.000  \\
\midrule 
~& BIL& BIR &  BOL &BOR & ELL &ELR&EUL&EUR&GTP&GTT&HP&HR& HY&NP& NR \\ 
\midrule 
$\mu$& 0.605& 0.655 & 0.613 &0.598  & 1.110 & 1.060 &0.989 & 0.978 &{0.074} &{0.045} &{0.004} &{0.005} & {-0.002} &{0.008} &{0.002}   \\ 
$\sigma$& 0.236 &0.233 & 0.211 & 0.216 & 0.521 &0.518 &0.311 &0.331 &{0.311}  &{0.113} &{0.036} & {0.021} & {0.048} &{0.024}&{0.009}\\
$V_{\text{max}}$& 1.000 & 1.000 & 1.000 & 1.000 & 2.000 &2.000 & 2.000 & 2.000 &{2.269}  & {1.082}&{0.328} &{0.300} &{0.355} & {0.204} &{0.084}\\
$V_{\text{min}}$&0.000 &0.000&0.000&0.000 &-1.000 &-1.000 &-1.000 &-1.000 &{-2.269} &{-0.826} &{-0.371}&{-0.173} &{-0.413} & {-0.104} &{-0.158}  \\
${V}_{\text{neu}}$&0.500 &0.500 &0.500 &0.500 &1.000  &1.000 &1.000 &1.000 &0.000 &0.000 &0.000 & 0.000 & 0.000 & 0.000 & 0.000\\
\bottomrule 
\end{tabular}
\end{adjustbox}
\label{tab:ctrl_stat}
\end{table*}

\begin{figure}
    \centering
\includegraphics[width=0.95\linewidth]{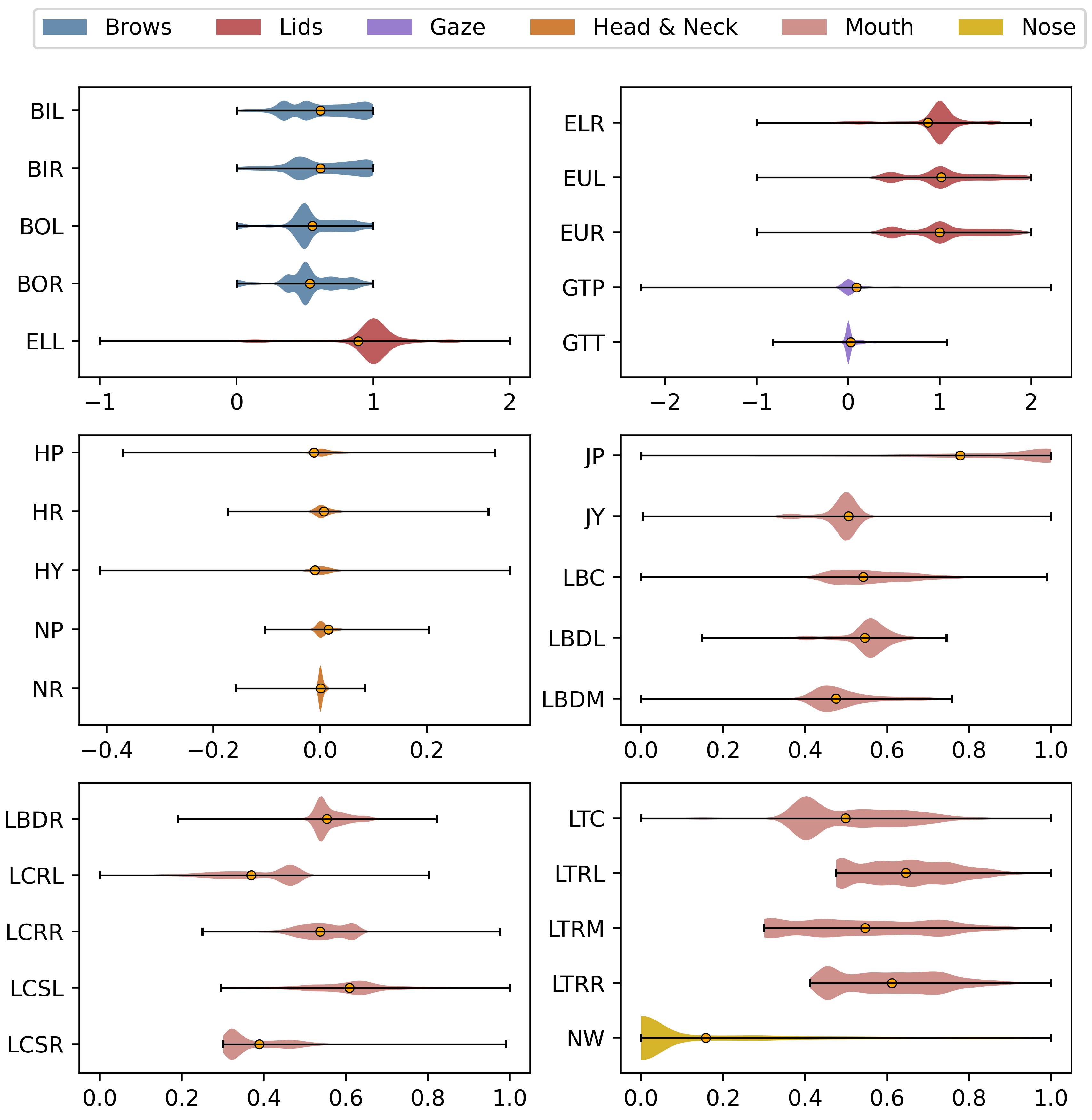}
    \caption{\textbf{Value distributions of 30 controls.} Controls for different expression-relevant units are indicated by different colors. The naming convention for controls is the same as in Table~\ref{tab:ctrl_range}. }
    \label{fig:ctrl-distribution}
\end{figure}
Examples from \textbf{X2C} are provided in Figure~\ref{fig:virtual_physical}, where the humanoid robot displays a wide range of facial expressions. Some expressions may not fit neatly into basic emotion categories (e.g., the 4th), some of them can be categorized into the same emotion class (fear) but with different intensities (e.g., the last two) and some of them contain asymmetric facial units (e.g., eyelids of the 2nd). We name the dataset \textbf{X2C} (Anything to Control) because it consists of pairs of emotion-aware representations (humanoid facial expressions, in this case) and their corresponding control values (used for guiding on-robot execution). We aim to expand this dataset by incorporating fine-grained emotion labels in our future work.
\paragraph{Characteristics and Quality}
Comparing to existing datasets~\cite{ chen2021smile,hu2024human} (Table~\ref{tab:benchmark-stat}) for humanoid facial expression imitation, \textbf{X2C} offers a larger scale and higher-dimensional annotations, enabling finer-grained robot control. Unlike prior works that rely on tools like MediaPipe~\cite{lugaresi2019mediapipe} to estimate facial landmarks---introducing potential errors---our control values are analytically calculated via interpolation functions, ensuring precise annotations that are perfectly aligned with images at each timestamp.
Distinctly, our dataset includes asymmetric facial expressions~\cite{kowner1995laterality}, which are closer to natural human behavior. This not only enhances diversity but also improves the expressive capacity of the robot.
We report summary statistics for each of the 30 control values in Table~\ref{tab:ctrl_stat}, including the mean ($\mu$), standard deviations ($\sigma$), minimum ($V_{\text{min}}$) and maximum ($V_{\text{max}}$).
Values in the neutral state ($V_{\text{neu}}$) 
 are also provided for reference. 
As shown, there are noticeable deviations between $\mu$ and $V_{\text{neu}}$ across most controls, and for many controls (e.g., JP, JY, LBC), the dataset samples span nearly the full achievable range as indicated in Table~\ref{tab:ctrl_range}, suggesting high-diversity. Note that we intentionally avoid sampling extreme values for certain controls because: 1) They could cause irreversible damage to the robot (e.g., excessive head or neck movement such as HP, HR, HY, NP, NR may lead to mechanical wear), and 2) Such expressions are physiologically implausible for humans---for instance, humans typically cannot fully hide their irises (GTP, GTT) or shape their lips into extreme forms like a `W' or `V' (see Figure~\ref{fig:weird}).
For clarity, a visualization of the value distributions for all 30 controls is provided in Figure~\ref{fig:ctrl-distribution}. 

\section{The X2CNet Framework}
\label{sec:framework}
\begin{figure}[h]
    \centering
    \includegraphics[width=0.95\linewidth]{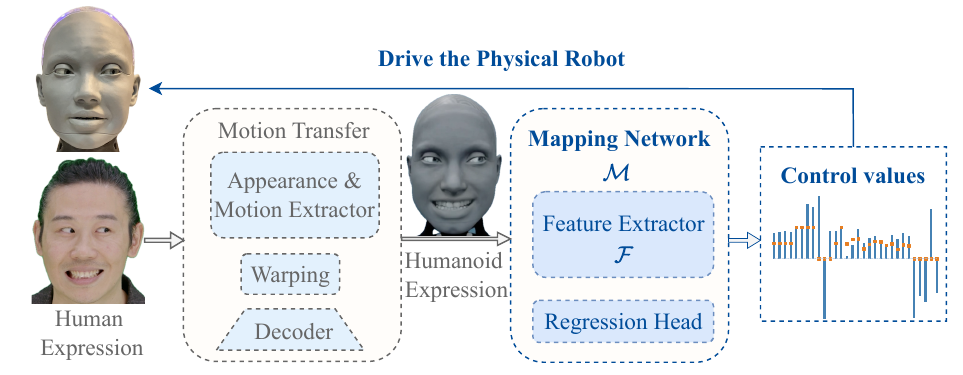}
    \caption{\textbf{An overview of X2CNet, the proposed imitation framework.} The first module captures facial expression subtleties from humans, while the mapping network learns the correspondence between various humanoid expressions and their underlying control values using the \textbf{X2C} dataset.}
    \label{fig:framework}
\end{figure}
\subsection{Motivation and Design} 
\label{sec:model-design}
The objective of this framework is to demonstrate the value of our dataset in advancing nuanced humanoid facial expression imitation. It should be able to learn from \textbf{X2C} and enable the humanoid robot to realistically mimic human facial expressions.
There are two key requirements for such a framework: 1) It must be capable of capturing subtle expression dynamics from humans. 2) It must output low-level action commands that interface with the robot's control system~\cite{fu2024humanplus,huang2016anticipatory, spong2020robot}. To meet the first requirement, we employ a motion transfer technique~\cite{siarohin2019first, siarohin2019animating, mallya2022implicit, guo2024liveportrait,wang2021one} to warp the humanoid face according to human expressions with emotional nuances in image space. To satisfy the second requirement, we learn the correspondence between nuanced humanoid facial expressions and the underlying control values using a mapping network trained on the \textbf{X2C} dataset.

\textbf{X2CNet} is thus composed of two modules (Figure~\ref{fig:framework}) and outputs 30 continuous control values that encode subtle movements of expression-relevant control units. Fine-grained control, together with a delicately designed humanoid face featuring 32 DoFs, makes realistic facial expression imitation possible.
From an implementation perspective, we adopt LivePortrait~\cite{guo2024liveportrait} as the motion transfer module, pretrained on a large corpus of high quality portrait data~\cite{nagrani2017voxceleb, wang2020mead,livingstone2018ryerson,liu2021blendgan}. As shown in Figure~\ref{fig:framework},
it consists of an appearance extractor, a motion extractor, a warping module, and a generator. The generated humanoid face---now expressing the human emotion---is then fed into a mapping network (denoted by $\mathcal{M}$), which consists of a feature extractor (denoted by $\mathcal{F}$) and a regression head. $\mathcal{F}$ is implemented using a ResNet18 backbone~\cite{he2016identity} while the regression head is a multilayer perceptron with two hidden layers and ReLU activations~\cite{jang2022bc}. 

\subsection{Experiments on X2C}
\label{sec:exp-on-x2c}
We split \textbf{X2C} into training and test sets, using 80\% of them for training. 
We use AdamW as the optimizer with a weight decay of 0.05 and apply a \textit{cosine schedule with warmup} as the learning rate scheduler, with an initial learning rate of 1e-3. The model is trained using the Huber loss with a threshold value of $\delta = 0.01$.
Throughout training, the batch size is set to 128, and the model is trained for 100 epochs.
All experiments are conducted on a single RTX 4090 GPU.

Control value prediction errors are evaluated on the test set using mean absolute error (MAE) as the performance measure.
To assess the effectiveness of the mapping network, we compare our method against three baselines. The first baseline randomly samples each control value independently from a uniform distribution (RC). The second baseline randomly selects samples directly from the training set (RT)~\cite{ho2016generative, cohen2013statistical}. While both involve random selection, they follow different strategies. The third baseline adopts the model architecture from~\cite{hu2024human}, predicting control values based on facial landmarks (LMKC). 
Quantitative results are summarized in Table~\ref{tab:comparison-result}. As shown, our method outperforms all three baselines on the test set consisting of 20,000 samples, achieving lower mean errors and smaller standard deviations.

Ablation studies are conducted on the feature extractor $\mathcal{F}$ within the mapping network, The MAEs for control value prediction, along with corresponding statistical analyses are reported in Table~\ref{tab:ablation}. 
Among the three CNN-based feature extractors---EfficientNet-B0, ResNet18, and VGG16---VGG16 achieves the best performance, followed by the transformer-based ViT-B/16. While ResNet18 performs slightly worse than both, it is significantly more lightweight and computationally efficient.

\begin{minipage}{0.45\textwidth}
 We compute the the mean absolute error (MAE) between the predicted and ground-truth control values, and conduct a detailed statistical analysis, including the calculation of the standard deviation (SD), standard error of the mean (SEM), and 95\% Confidence Interval (CI).  
\end{minipage}
\hfill
\begin{minipage}{0.50\textwidth}
\centering
    \centering
    \captionof{table}{Comparison results based on MAE and corresponding statistical analysis.}
    \begin{adjustbox}{width=\columnwidth,center}
    \begin{tabular}{ccccc}
    \toprule
         Method& MAE$~\downarrow$ &SD$~\downarrow$ &SEM$~\downarrow$ &95\% CI\\
         \midrule
         RC& 0.8951 & 0.7217 & 0.0051 & $[0.8851, 0.9051]$\\
         RT &1.0629 & 0.8904 & 0.0063 & $[1.0505, 1.0752]$\\
         LMKC & 0.1602 & 0.3402 & 0.0024 & $[0.1555, 0.1629]$\\
         \textbf{OURS} &\textbf{0.0114} & \textbf{0.0650} & \textbf{0.0005} & $[0.0105, 0.0123]$\\
         \bottomrule
    \end{tabular}
    \label{tab:comparison-result}
    \end{adjustbox}
\end{minipage}

\begin{minipage}{0.45\textwidth}
 We study alternative structures of the feature extractor $\mathcal{F}$ within the mapping network by replacing it with the backbone of EfficientNet-B0 (EN-B0)~\cite{tan2019efficientnet},  VGG16~\cite{simonyan2014very} and Vision Transformer (ViT-B/16)~\cite{dosovitskiy2020image}. All of them are adapted from official implementations and initialized with pretrained weights.
\end{minipage}
\hfill
\begin{minipage}{0.50\textwidth}
\centering
    \centering
    \captionof{table}{An ablation study on the feature extractor.} 
    \begin{adjustbox}{width=\columnwidth,center}
    \begin{tabular}{ccccc}
    \toprule
         $\mathcal{F}$& MAE$~\downarrow$ &SD$~\downarrow$ &SEM$~\downarrow$ &95\%CI \\
         \midrule
         EN-B0& 0.0151 & 0.0636 &0.0004 &$[0.0142, 0.0159]$\\
         VGG16 &{0.0107} & {0.0642} &0.0005 &$[0.0098, 0.0116]$\\
         ViT-B/16 & 0.0111 & 0.0641 &0.0005 &$[0.0103, 0.0120]$\\
         ResNet18 &{0.0114} & {0.0650} & {0.0005} & $[0.0105, 0.0123]$\\
         \bottomrule

    \end{tabular}
    \label{tab:ablation}
    \end{adjustbox}
\end{minipage}






\section{Real-World Demonstrations}
\label{sec:realworld-demo}
\begin{figure}
    \centering
    \includegraphics[width=1.0\linewidth]{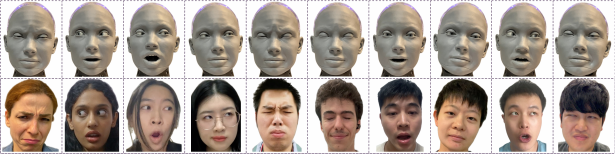}
    \caption{\textbf{Examples of realistic humanoid imitation.} Different individuals express a wide range of facial expressions, with nuances
reflected in features such as frown, gaze direction, eye openness, nose wrinkles, mouth openness, and so on. These nuanced human facial
expressions extend beyond canonical emotions and can be regarded as either blends of different canonical emotions or as a single emotion
with varying intensities. The humanoid robot mimics every detail, resulting in a realistic imitation.}
\label{fig:realworld-demos}
\end{figure}
To meet the requirements of real-world demonstrations, the robot must be capable of imitating a wide range of nuanced facial expressions from diverse human performers in the wild.
To this end, we particularly recruited 20 human performers from 5 different countries, including both males and females. As shown in Figure~\ref{fig:realworld-demos}, They exhibit a variety of facial contours, skin tones, and hairstyles. Their facial expressions were captured under different lighting conditions, and some performers appear with accessories such as glasses (4th) and earphones (6th). 
Performers were instructed to go beyond canonical expressions by incorporating various subtleties such as frowning (1st), gaze direction (2nd), and neck movement (8th). As illustrated in Figure~\ref{fig:realworld-demos}, the humanoid robot successfully mimics most of these nuanced expressions despite its hardware constraints. These results validate the effectiveness of the imitation framework and, more importantly, highlight the value of \textbf{X2C}---our high-diversity, high-quality, large-scale dataset for humanoid learning.
More real-world demonstrations can be found on our project website: \href{https://lipzh5.github.io/X2CNet/}{https://lipzh5.github.io/X2CNet/}.




\section{Related Work}
\paragraph{Facial Expressions for Affective Human-Robot Interaction} Facial expressions has been proved an indispensable mode of affective communication~\cite{mehrabian2017communication, rawal2022facial} and numerous studies 
have examined the important role of facial expressions in affective human-robot  interaction (HRI)~\cite{de2016long,nicolescu2001learning,ray2008people,li2024ugotme, cao2024ai}. 
However, some of them focus on human facial expression analysis, neglecting the emotionally intelligent behavior on robot's face, where the robots may only display limited categories of emotional signals (such as LED indicators) on their faces~\cite{barros2015emotional,liu2017facial,johnson2013imitating, faria2017affective}. The robots often fail to convey the nuances of emotions, leading to reduced user engagement and trust in HRI. 
By providing a large-scale, annotated humanoid facial expression dataset, \textbf{X2C} pave the path for researchers aiming at improving the robots' facial expressiveness for HRI. 
\paragraph{Humanoid Facial Expressions Imitation} Although recent studies have paid more attention to the expression display on robots' faces~\cite{meghdari2016real,cid2013imitation, ge2008facial,rawal2022exgennet}, only a limited set of facial expressions are covered, which limits the expressiveness of the robot.
Although efforts have been paid on nuanced facial expression imitation recently~\cite{antony2025xpress, hu2024human,chen2021smile, liu2024unlocking,li2023design}, there is a lack of public available resources for accessing advanced, delicate humanoid face, benchmarking different models on this task.  
To bridge the gap, we introduce \textbf{X2C}, the first high-quality, high-diversity, large-scale dataset featuring nuanced humanoid facial expressions with precise control value annotations for realistic humanoid imitation.

\section{Limitations, Discussions and Conclusions}
\label{sec:conclusion}
\paragraph{Limitations and Future Work} 
While our dataset collection includes volunteers from multiple countries and genders, cultural biases may still be present, potentially influencing the interpretation or design of facial expressions. In the future, we plan to expand the sample population for recruiting animation creators and to further diversify the dataset.
Although our current dataset includes facial expressions from only a single humanoid robot, the data collection pipeline and the proposed imitation framework are designed to generalize to other humanoid platforms with different degrees of freedom (DoFs). Future work will also focus on extending \textbf{X2C} with fine-grained emotion labels to enable more precise supervision for the imitation task.

\paragraph{Ethical Considerations and Societal Impacts} All human participants involved in the real-world experiments provided informed consent, with consent forms included in the Supplemental Material. No harmful information about participants is released, and all data use follows ethical research guidelines.
The dataset could be misused for deceptive, manipulative, or surveillance-related purposes, such as impersonation or unauthorized identity mimicry. We strongly discourage such applications and advocate for responsible, ethical AI use.
Positively, the dataset has the potential to empower emotionally intelligent robots for socially beneficial applications, including elderly care, autism therapy, and education, by enabling more expressive and relatable interactions. However, overly human-like robots may cause users—especially vulnerable individuals—to form emotional attachments or unrealistic expectations, possibly leading to confusion or psychological discomfort. These risks must be carefully managed through transparent system design and user education.

\paragraph{Conclusions} 
To equip the humanoid robot with the ability to realistically imitate nuanced human facial expressions, we make the following contributions: 1) we introduce the \textbf{X2C} (Anything to Control)---a high-quality, high-diversity, large-scale datasets featuring nuanced humanoid facial expressions with precise control value annotations; 2) we propose \textbf{X2CNet}, a novel framework for human-to-humanoid expression imitation; 3) we provide real-world demonstrations on the physical robot to validate the effectiveness of our method, and the potential of our dataset in advancing realistic humanoid facial expression imitation. 
 


\bibliographystyle{unsrt}
\bibliography{mybib}

\appendix
\section*{Appendix}
\label{sec:appendix}
\renewcommand{\thefigure}{A\arabic{figure}}
\setcounter{figure}{0}


\begin{minipage}{0.50\textwidth}
    \centering
\includegraphics[width=1.0\linewidth]{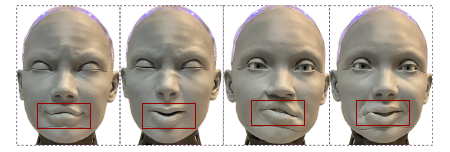}
    \captionof{figure}{Examples of facial expressions that are physiologically implausible for humans (`W' shape, `V' shape, and two asymmetric cases).}
    \label{fig:weird}
\end{minipage}
\hfill
\begin{minipage}{0.45\textwidth}
 Some combinations of control values may lead to facial expressions that are physically implausible for humans. We present several examples in Figure~\ref{fig:weird} with a focus on the mouth. Both symmetric (left two) and asymmetric (right two) cases are provided.
\end{minipage}

\begin{figure}[h]
    \centering
    \includegraphics[width=0.40\linewidth]{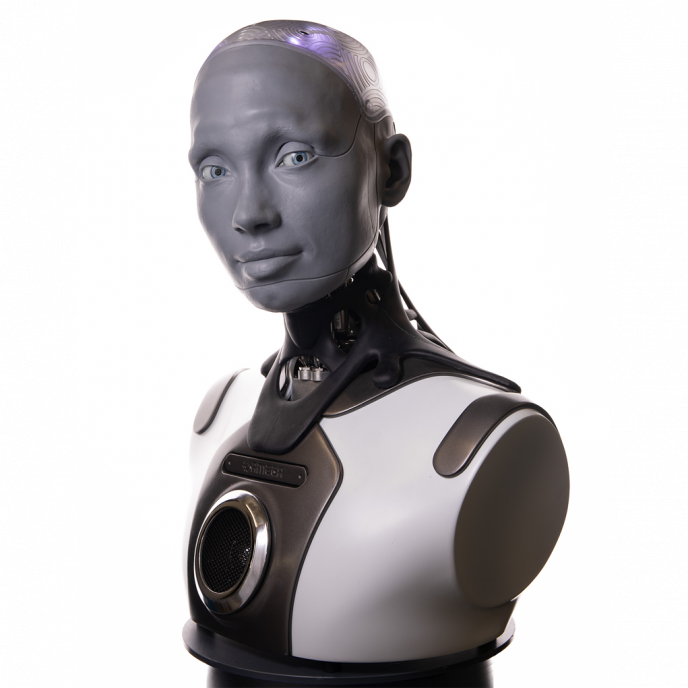}
    \caption{An image of the humanoid robot used for dataset collection and experiments. (Image source: \href{https://engineeredarts.com/robot/ameca/}{https://engineeredarts.com/robot/ameca/}).}
    \label{fig:ameca}
\end{figure}


\begin{figure}[h]
    \centering
    \includegraphics[width=1.0\linewidth]{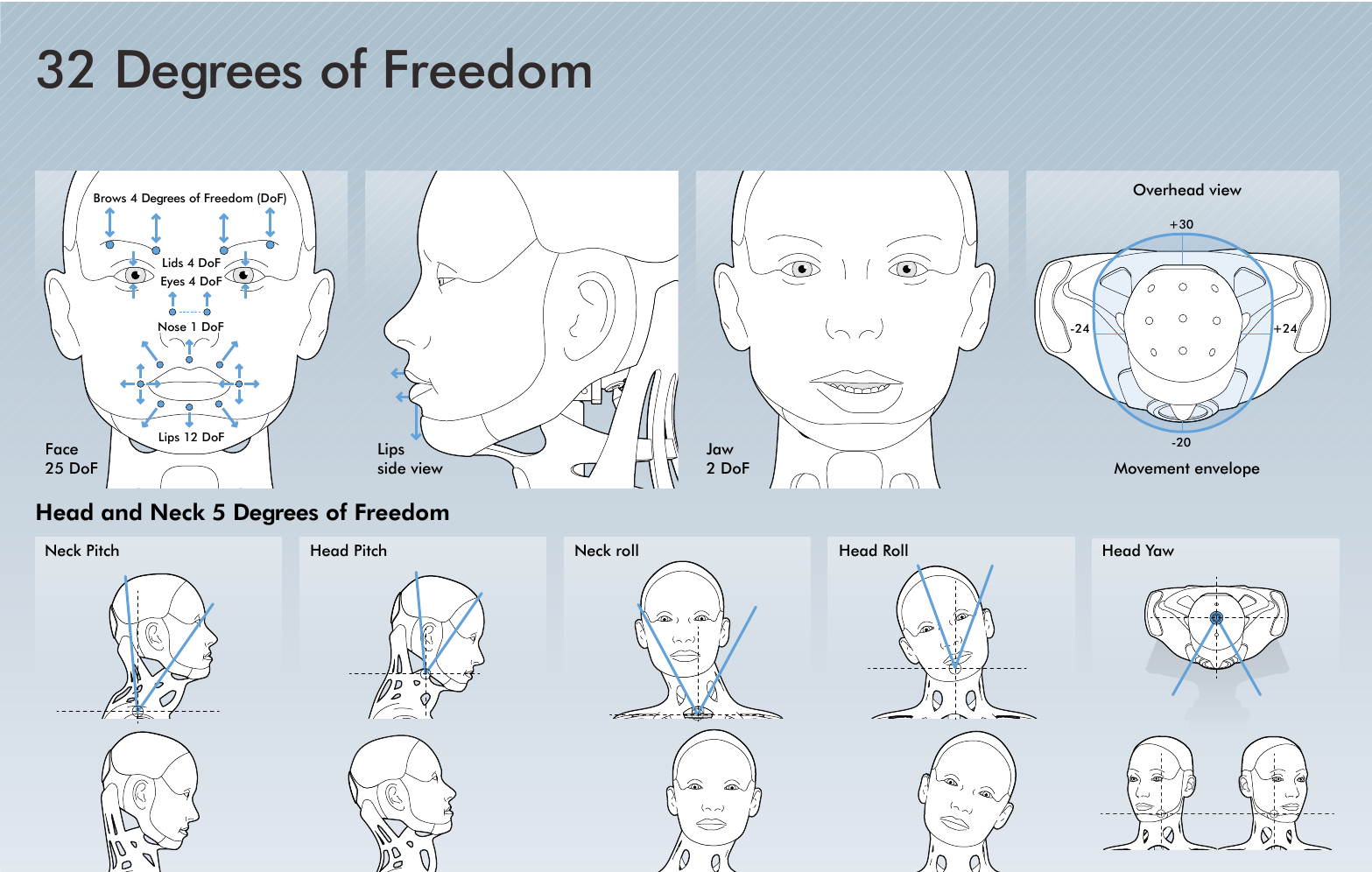}
    \caption{Demonstration of the 32 Degrees of Freedom (DoFs) on the robot’s face (Figure source: \href{https://engineeredarts.com/robot/ameca/}{https://engineeredarts.com/robot/ameca/}).}
    \label{fig:32DoF}
\end{figure}





\newpage

\end{document}